\documentclass[twoside,11pt]{article}

\usepackage[abbrvbib, preprint]{jmlr2e}

\usepackage{lastpage}
\usepackage{tabularx}
\usepackage{amsmath}
\usepackage{amsfonts}
\usepackage{booktabs}
\usepackage{longtable}
\usepackage{listings}

\PassOptionsToPackage{table}{xcolor}
\usepackage{xcolor}
\usepackage{booktabs}
\usepackage{multirow}
\usepackage{xcolor}

\usepackage{dirtree}

\usepackage{hyperref}
\usepackage{colortbl}
\usepackage[most]{tcolorbox}

\usepackage{booktabs}
\usepackage[table]{xcolor}
\usepackage{pifont}
\usepackage{colortbl}
\usepackage{graphicx}

\usepackage[most]{tcolorbox}
\usepackage{tabularx}
\usepackage{booktabs}
\usepackage{array}
\usepackage{enumitem}
\usepackage[table]{xcolor}
\usepackage{tikz}
\usetikzlibrary{shadows.blur, positioning}

\usepackage[dvipsnames]{xcolor}
\usepackage[most]{tcolorbox}
\tcbuselibrary{skins,breakable,listings}

\usepackage{listings}

\usepackage{enumitem}

\definecolor{MainBlue}{RGB}{32,74,135}
\definecolor{SoftBlue}{RGB}{239,245,255}
\definecolor{LineBlue}{RGB}{88,126,191}
\definecolor{AccentBlue}{RGB}{15,52,96}

\newcommand{\cmark}{\textcolor{green!70!black}{\ding{51}}}
\newcommand{\xmark}{\textcolor{red!75!black}{\ding{55}}}

\tcbset{
    colback=blue!5,
    colframe=blue!60!black,
    fonttitle=\bfseries,
    coltitle=black
}

\newtcolorbox{scalabilitynote}{
  colback=green!5,
  colframe=green!50!black,
  colbacktitle=green!15,
  coltitle=black,
  fonttitle=\bfseries,
  title=Scalability Note
}

\newcommand{\pmark}{\textcolor{orange!80!black}{$\circ$}}

\definecolor{codegreen}{rgb}{0,0.6,0}
\definecolor{codegray}{rgb}{0.5,0.5,0.5}
\definecolor{codepurple}{rgb}{0.58,0,0.82}
\definecolor{backcolour}{rgb}{0.95,0.95,0.92}

\lstdefinestyle{mystyle}{
    backgroundcolor=\color{backcolour},   
    commentstyle=\color{codegreen},
    keywordstyle=\color{magenta},
    numberstyle=\tiny\color{codegray},
    stringstyle=\color{codepurple},
    basicstyle=\ttfamily\footnotesize,
    breakatwhitespace=false,         
    breaklines=true,                 
    captionpos=b,                    
    keepspaces=true,                 
    numbers=left,                    
    numbersep=5pt,                  
    showspaces=false,                
    showstringspaces=false,
    showtabs=false,                  
    tabsize=2
}

\lstdefinestyle{pythonstyle}{
    language=Python,
    basicstyle=\ttfamily\small,
    keywordstyle=\color{blue},
    commentstyle=\color{ForestGreen},
    stringstyle=\color{BrickRed},
    backgroundcolor=\color{gray!8},
    frame=single,
    rulecolor=\color{black!20},
    breaklines=true,
    showstringspaces=false,
    tabsize=4
}

\lstdefinestyle{bashstyle}{
    language=bash,
    basicstyle=\ttfamily\small,
    backgroundcolor=\color{gray!8},
    frame=single,
    rulecolor=\color{black!20},
    breaklines=true,
    showstringspaces=false
}

\newtcolorbox{testingstats}{
    enhanced,
    breakable,
    colback=blue!3,
    colframe=blue!70!black,
    coltitle=black,
    title=Testing Statistics,
    fonttitle=\bfseries,
    boxrule=0.8pt,
    arc=2mm,
    left=3mm,
    right=3mm,
    top=2mm,
    bottom=2mm,
    drop shadow
}

\makeatletter
\renewcommand\paragraph{%
  \@startsection{paragraph}{4}{\parindent}%
    {2.25ex \@plus1ex \@minus.2ex}%
    {-1em}%
    {\normalfont\normalsize\itshape}}
\makeatother

\jmlrheading{1}{2026}{1-\pageref{LastPage}}{X/26}{X/26}{XX-XXXX}{Kiyan Rezaee, Morteza Ziabakhsh, Artin Bahrampour, Seyed Mohammad Ghoreishi, Asal Khajeh, Ali Sajedifar, Manny Chalak, Ava Zerafatangiz and Sadegh Eskandari}
\ShortHeadings{SCPP: A Python Library for Soft Clustering}{Rezaee et al.}
\firstpageno{1}

\begin{document}

\title{SCPP: A Unified Python Library for Soft Clustering}

\author{%
\name Kiyan Rezaee \email kiyanrezaee17@gmail.com \\
\name Morteza Ziabakhsh \email morteza24mail@protonmail.com \\
\name Artin Bahrampour \email Artinbahrampour1599@gmail.com \\
\name Seyed Mohammad Ghoreishi \email mohammad.ghoreishi530@gmail.com \\
\name Asal Khajeh \email Asalkhajeh1@gmail.com \\
\name Ali Sajedifar \email Its.AliSajedifar@gmail.com \\
\name Manny Chalak \email manee.ch83@gmail.com \\
\name Ava Zerafatangiz \email ava.zerafattt@gmail.com \\
\name Sadegh Eskandari \email eskandari@guilan.ac.ir  \\
\addr Department of Computer Science, University of Guilan, Rasht, Iran%
}

\editor{}

\maketitle

\begin{abstract}%
Soft clustering implementations are fragmented across Python packages with
inconsistent estimator interfaces, parameter conventions, and membership
representations, making systematic comparison across methodological families
difficult. We present \textbf{SCPP} (\textbf{\underline{S}oft
\underline{C}lustering \underline{P}ython \underline{P}ackage}), an
open-source Python library that unifies 42 soft clustering algorithms covering
fuzzy, probabilistic, graph-based,
matrix-factorization, and deep-learning approaches under a common
protocol. Every model exposes soft assignments
through a membership matrix $U \in \mathbb{R}^{n \times K}_{\geq 0}$, together
with labels, prototypes, and cluster-count information under consistent
semantics. A conformance suite validates this protocol for all 42 models
in continuous integration. SCPP also provides a unified evaluation framework
covering 20 data sets and 12 cluster-validity metrics, enabling cross-family
experiments under a common interface. Four
models are additionally compared against independently maintained
implementations of the same algorithm; where the two membership matrices
encode the same quantity, the largest disagreement is $6.2\times10^{-10}$, and
all hard partitions are identical. SCPP is distributed through PyPI under the MIT license and is supported by
1{,}206 automated tests across 48 modules, 95\% line coverage, and continuous
integration across three operating systems and four Python versions.
\end{abstract}


\section{Introduction}

Soft clustering assigns each observation a vector of memberships over clusters
rather than a single label. This representation is useful when observations
may participate in multiple groups: genes can belong to several biological
pathways, documents can cover multiple topics, network nodes can belong to
overlapping communities, and software classes can exhibit cross-cutting
responsibilities \citep{dogan2025fuzzy,liu2025llm,cangemi2025fuzzy,
chi2025novel,ziabakhsh2025extracting,rezaee2025fos}.

Despite the breadth of soft clustering methods, their Python implementations
remain fragmented across methodological families. Several packages provide
substantial coverage within particular families: \texttt{scikit-fuzzy}
\citep{scikit_fuzzy_2024}, \texttt{fuzzy-c-means} \citep{dias2019fuzzy}, and
\texttt{fuzzycat} \citep{oliver_fuzzycat} support fuzzy clustering;
\texttt{scikit-learn} \citep{pedregosa2011scikit} and
\texttt{Mixture-Models} \citep{kasa2024mixture} provide probabilistic mixture
models; \texttt{ClustPy} \citep{leiber2023benchmarking} covers deep clustering;
and graph libraries such as \texttt{CDlib} \citep{rossetti2019cdlib} and
\texttt{karateclub} \citep{rozemberczki2020karateclub} provide common
interfaces for collections of community-detection methods. These packages
demonstrate that unification within a methodological family is practical.
The remaining gap is cross-family support: a study comparing possibilistic, evidential, probabilistic, graph-based, and deep soft clustering methods must currently combine multiple libraries, reconcile their conventions, and adapt their outputs to a common evaluation workflow.

SCPP (\textbf{\underline{S}oft \underline{C}lustering
\underline{P}ython \underline{P}ackage}) addresses this integration problem
without imposing a common input representation. Different algorithms
necessarily require different inputs,  for example, feature matrices,
graphs, or collections of partitions, and SCPP preserves those
algorithm-specific requirements. Instead, it standardizes the interface and
semantics of the fitted results: every estimator exposes a membership matrix
$U \in \mathbb{R}^{n \times K}_{\geq 0}$ together with hard labels, cluster
prototypes where applicable, and cluster-count information. This separation
allows heterogeneous algorithms to retain their native computational
requirements while making their outputs accessible to common testing,
evaluation, and benchmarking infrastructure.

Our contributions are as follows:

\begin{enumerate}[leftmargin=*,topsep=2pt,itemsep=1pt]

\item \textbf{A common estimator protocol.}
All SCPP estimators implement \texttt{fit} and expose standardized attributes, \texttt{memberships\_}, \texttt{labels\_}, \texttt{centers\_}, and
\texttt{n\_clusters}, with defined shapes and semantics. A conformance
suite constructs and fits all 42 exported estimators and verifies the
protocol in continuous integration (Section~\ref{sec:design}, Section~\ref{sec:quality}).

\item \textbf{Cross-family coverage.}
SCPP provides 42 algorithms spanning fuzzy, possibilistic, evidential,
probabilistic, graph-based, matrix-factorization, document, ensemble, and
deep soft clustering (Table~\ref{tab:comparison},
Appendix~\ref{app:algorithms}).

\item \textbf{A unified evaluation and benchmarking framework.}
SCPP provides 20 data sets, 12 cluster-validity metrics, and tools for runtime,
memory, and scalability measurement. We use this infrastructure to evaluate
all 20 feature-matrix estimators across the eleven data sets, yielding
220 fits under a common protocol. (Section~\ref{sec:benchmark}, Appendix~\ref{app:performance}).

\item \textbf{Reproducible software infrastructure.}
The library includes 1{,}206 automated tests across 48 modules, 95\% line
coverage, continuous integration across three operating systems and four
Python versions, 42 runnable examples, and distribution through PyPI under
the MIT license (Section~\ref{sec:quality}).

\end{enumerate}

\begin{table}[t]
\centering
\footnotesize
\setlength{\tabcolsep}{4pt}
\definecolor{headerbg}{RGB}{232,245,233}
\definecolor{scppbg}{RGB}{220,252,231}
\definecolor{altrow}{RGB}{248,250,252}
\renewcommand{\arraystretch}{1.15}

\begin{tabular}{lrcccccc}
\toprule
\textbf{Package}
& \textbf{Soft}
& \textbf{Fuzzy}
& \textbf{Prob.}
& \textbf{Graph}
& \textbf{Deep}
& \textbf{Checked}
& \textbf{Benchmarking} \\
& \textbf{methods}
&
&
&
&
& \textbf{API}
& \\
\midrule

\texttt{fuzzy-c-means}~\citep{dias2019fuzzy}
& 1 & \cmark & \xmark & \xmark & \xmark & \xmark & \xmark \\

\texttt{fuzzycat}~\citep{oliver_fuzzycat}
& 1 & \cmark & \xmark & \xmark & \xmark & \xmark & \xmark \\

\texttt{scikit-fuzzy}~\citep{scikit_fuzzy_2024}
& 1 & \cmark & \xmark & \xmark & \xmark & \xmark & \xmark \\

\texttt{HDBSCAN}~\citep{mcinnes2017hdbscan}
& 1 & \xmark & \pmark & \xmark & \xmark & \xmark & \xmark \\

\texttt{pyclustering}~\citep{novikov2019pyclustering}
& 2 & \cmark & \pmark & \xmark & \xmark & \xmark & \xmark \\

\texttt{scikit-learn}~\citep{pedregosa2011scikit}
& 4 & \xmark & \cmark & \xmark & \xmark & \cmark & \pmark \\

\texttt{ClustPy}~\citep{leiber2023benchmarking}
& 6 & \pmark & \cmark & \xmark & \cmark & \xmark & \cmark \\

\texttt{Mixture-Models}~\citep{kasa2024mixture}
& 8 & \xmark & \cmark & \xmark & \xmark & \xmark & \xmark \\

\texttt{karateclub}~\citep{rozemberczki2020karateclub}
& 5 & \xmark & \xmark & \cmark & \pmark & \xmark & \xmark \\

\texttt{CDlib}~\citep{rossetti2019cdlib}
& 15 & \pmark & \xmark & \cmark & \pmark & \xmark & \cmark \\

\midrule
\rowcolor{scppbg}
\textbf{SCPP (this work)}
& \textbf{42}
& \cmark
& \cmark
& \cmark
& \cmark
& \cmark
& \cmark \\

\bottomrule
\end{tabular}
\caption{Python packages providing soft clustering methods.
\emph{Soft methods} counts estimators that expose a per-sample membership,
responsibility or affiliation matrix through the package's public API. Counts were taken from
each package's public API in August 2026.
\cmark~= supported, \pmark~= partial, \xmark~= not provided.}
\label{tab:comparison}
\end{table}

\section{Design and Implementation}
\label{sec:design}

\paragraph{What is standardized, and what is not.}
SCPP does not force different algorithms to accept the same input. Prototype
methods take a feature matrix, graph methods an adjacency or affinity matrix,
consensus methods a set of partitions, document methods raw text; the first
argument of \texttt{fit} is whatever the algorithm actually needs.
What SCPP standardizes is the fitted state. Every estimator accepts its
parameters through \texttt{\_\_init\_\_}, \texttt{fit} returns \texttt{self},
and \texttt{memberships\_} is an $(n, K)$ matrix with non-negative entries;
\texttt{labels\_} is its row-wise $\arg\max$, and \texttt{centers\_} is
provided when the method has prototypes.

\paragraph{Implementation.}
Each algorithm lives in its own module
under \texttt{soft\_clustering/} with a reference to its original
publication. Estimators are imported lazily, so \texttt{import
soft\_clust ering} loads no estimator module; the core
installation is NumPy, SciPy, scikit-learn and \texttt{typeguard}. The four
PyTorch-backed models (CDCGS, DMoN, NOCD, RDFKC) sit behind the \texttt{deep}
extra and raise an explicit installation message when it is absent. A CI job
installs the built wheel into a clean environment and imports and constructs
every estimator. A minimal example of fitting an estimator and accessing its
standardized outputs is shown below.

\begin{lstlisting}[language=Python, caption={Fitting an estimator (FCM) and reading its soft output.}]
from sklearn.datasets import load_iris
from soft_clustering import FCM

X, y = load_iris(return_X_y=True)
model = FCM(n_clusters=3, m=2.0, random_state=0).fit(X)
U = model.memberships_        # (150, 3), rows sum to 1
centers = model.centers_      # (3, 4)
labels = model.labels_        # (150,), argmax over U
\end{lstlisting}


\paragraph{Extension.}
Adding an algorithm to SCPP requires only implementing its training procedure and exposing the membership matrix; the shared infrastructure handles the rest. \texttt{BaseSoftClusterer} derives hard labels, prediction methods, and cluster counts from the membership matrix and normalizes differing constructor conventions for $K$. The developer then registers the estimator, adds a conformance entry, and provides its documentation and runnable example. Once registered, the estimator automatically participates in benchmarking and evaluation without adapters or estimator-specific integration code.

\section{Evaluation and Benchmarking}
\label{sec:benchmark}

\paragraph{Data sets.} \texttt{benchmarking.datasets} returns a standardised
$(X, y)$ pair for 20 data sets in three groups: five real (iris, wine, digits,
breast cancer, olivetti faces), six synthetic generators with controlled
overlap and anisotropy, and nine OpenML data sets ($n$ from 214 to 20{,}000,
$K$ from 4 to 26). Ground-truth labels are used for external validation.

\paragraph{Metrics.} \texttt{soft\_clustering\_metrics} computes seven measures
on $U$, partition coefficient, modified partition coefficient, partition
entropy, fuzzy hypervolume, Xie--Beni index, fuzzy compactness, and fuzzy
separation, and \texttt{clustering\_metrics} computes five on defuzzified
labels (silhouette, Calinski--Harabasz, Davies--Bouldin, ARI, NMI). Entropies
use the natural logarithm throughout.

\paragraph{Computational measurement.}
\texttt{RuntimeBenchmark}, \texttt{MemoryBenchmark}, and
\texttt{Scalability Benchmark} measure fit and prediction time, memory
consumption, and their scaling with data set size. \texttt{ClusteringBenchmark}
runs a selected set of estimators across a selected set of data sets and returns
the results as a \texttt{pandas} DataFrame. Appendix~\ref{app:performance} reports the cross-family comparison: all 20
estimators that consume a feature matrix, over 11 data sets, 220
fits under one protocol. Appendix~\ref{app:external} reports agreement and
runtime against third-party implementations of the same algorithms.
\section{Software Quality, Documentation, and Availability}
\label{sec:quality}

\paragraph{Testing.}
The test suite contains 1{,}206 tests across 48 modules with 95\% line
coverage. The \emph{conformance suite} constructs and fits all 42
exported estimators and verifies that \texttt{fit} returns \texttt{self},
\texttt{memberships\_} has shape $(n,K)$ and valid values, partition-constrained
memberships sum to one, \texttt{labels\_} equals $\arg\max U$, prediction
before fitting fails, \texttt{predict(X)} does not silently return training
labels, parameters round-trip through \texttt{sklearn.clone}, and, for the
estimators that accept one, a fixed \texttt{random\_state} reproduces the
memberships. The suite is tied to the
public registry with no exclusions, so every exported estimator must satisfy
the contract to pass CI.

\paragraph{Continuous integration.} GitHub Actions runs the suite on Ubuntu,
macOS and Windows for Python 3.10--3.13.
Further jobs install the package into a clean environment and import and
construct every estimator, then fit one estimator from each input modality;
build and inspect the source and wheel distributions; run the benchmarking
suite against its optional extra; execute all 42 example
scripts; and run \texttt{ruff} and \texttt{black}.

\paragraph{Documentation.} Installation instructions, per-algorithm API
reference and runnable examples are published at
\url{https://soft-clustering.readthedocs.io} and rebuilt in CI. The
repository carries 42 standalone example scripts, one per algorithm, a
contributing guide, a code of conduct, a changelog, and issue templates.

\paragraph{Availability.} SCPP is released under the MIT license, developed
in the open at \url{https://github.com/soft-clustering/soft-clustering}, and
installed with \texttt{pip install soft- clustering} (or
\texttt{soft-clustering[deep]} for the PyTorch models,
\texttt{soft-clustering[bench]} to run the benchmarking suite). The
cross-family comparison, the third-party agreement checks and the
optimization study are regenerated from the repository by the commands listed
in Appendices~\ref{app:performance}, \ref{app:external} and
\ref{app:optimization}.




\newpage

\appendix


\begin{tcolorbox}[
    enhanced,
    breakable,
    colback=white,
    colframe=MainBlue,
    coltitle=white,
    colbacktitle=MainBlue,
    fonttitle=\bfseries\Large,
    title={Appendix Overview},
    arc=4mm,
    boxrule=1pt,
    left=6mm,
    right=6mm,
    top=5mm,
    bottom=5mm,
    attach boxed title to top center={yshift=-2mm},
    boxed title style={
        enhanced,
        arc=3mm,
        boxrule=0pt,
        colback=MainBlue,
    },
    drop fuzzy shadow,
    before skip=12pt,
    after skip=12pt
]


\begin{center}
{\Large\bfseries\textcolor{AccentBlue}{
Supplementary Material for the SCPP}}
\end{center}

\vspace{0.8em}


\renewcommand{\arraystretch}{1.7}

\begin{tabularx}{\textwidth}{
    >{\centering\arraybackslash\bfseries\color{MainBlue}}m{1.2cm}
    >{\raggedright\arraybackslash}X
    >{\centering\arraybackslash\bfseries\color{MainBlue}}m{2cm}
}

\toprule

\rowcolor{SoftBlue}

\Large\textbf{Sec.}
&
\centering\arraybackslash\Large\textbf{Content}
&
\Large\textbf{Page}
\\

\midrule

\rowcolor{gray!3}


\rowcolor{gray!3}

\Large\textbf{A}
&
\large\bfseries Implemented Algorithms in SCPP
&
\Large\pageref{app:algorithms}
\\


\rowcolor{blue!1}

\Large\textbf{B}
&
\large\bfseries Benchmark Data Set Collection
&
\Large\pageref{app:dataset}
\\


\rowcolor{gray!3}

\Large\textbf{C}
&
\large\bfseries Soft Clustering Evaluation Framework
&
\Large\pageref{app:metrics}
\\


\rowcolor{blue!1}

\Large\textbf{D}
&
\large\bfseries Cross-Family Comparison

&
\Large\pageref{app:performance}
\\

\rowcolor{gray!3} \Large\textbf{E} & \large\bfseries Agreement with Independent Implementations & \Large\pageref{app:external} \\

\rowcolor{blue!1} \Large\textbf{F} & \large\bfseries Testing & \Large\pageref{Testing} \\


\rowcolor{gray!3} \Large\textbf{G} & \large\bfseries Case Study: A Migration Pipeline & \Large\pageref{appendix:mo2om} \\


\rowcolor{blue!1} \Large\textbf{H} & \large\bfseries Implementation Optimization Study & \Large\pageref{app:optimization} \\



\bottomrule

\end{tabularx}

\vspace{1em}

\end{tcolorbox}

\pagebreak

\section{Implemented Algorithms in SCPP}
\label{app:algorithms}

Table~\ref{tab:scpp_algorithm_families} lists all algorithms currently
exported by SCPP together with their input type. Each row corresponds to one
implementation and one conformance entry; the test suite asserts that the
public registry and the conformance table name exactly the same estimators, in
both directions.

Three cases require clarification. \textbf{SCSPA, SHBGF, and SMCLA} produce
crisp partitions; SCPP represents these as one-hot membership matrices to
preserve a common downstream interface. \textbf{KMART} may produce
non-normalized memberships because its
underlying ART formulation allows samples to belong to multiple
or no categories. Finally, \textbf{LDA} produces document--topic
distributions, which SCPP treats as soft memberships; its topic--word
distribution remains available separately.

\begin{table*}[tp]
\centering
\footnotesize
\setlength{\tabcolsep}{3.5pt}
\renewcommand{\arraystretch}{1.05}

\begin{tabularx}{\textwidth}{@{}p{1.7cm} l X@{}}
\toprule
\textbf{Family} & \textbf{Estimator} & \textbf{Method and input} \\
\midrule

\rowcolor{gray!10}
Foundational & \texttt{FCM} & Fuzzy c-means~\cite{bezdek1981pattern}; features \\
\rowcolor{gray!10}
& \texttt{GMM} & Gaussian mixture EM~\cite{dempster1977maximum}; features \\
\rowcolor{gray!10}
& \texttt{PCM} & Possibilistic c-means~\cite{krishnapuram1993possibilistic}; features \\
\rowcolor{gray!10}
& \texttt{GK} & Gustafson--Kessel~\cite{gustafson1979fuzzy}; features \\
\rowcolor{gray!10}
& \texttt{GathGeva} & Gath--Geva FMLE~\cite{gath1989fuzzy}; features \\

Classical
& \texttt{RoughKMeans} & Rough $k$-means~\cite{lingras2004interval}; features \\
& \texttt{LDA} & Latent Dirichlet allocation~\cite{blei2003latent}; documents \\
& \texttt{PLSI} & Probabilistic LSA~\cite{hofmann1999plsa}; documents \\
& \texttt{SCM} & Subtractive clustering~\cite{chiu1994fuzzy}; features \\

\rowcolor{gray!10}
Kernel and
& \texttt{KFCM} & Kernelised FCM~\cite{zhang2004novel}; features \\
\rowcolor{gray!10}
density
& \texttt{KFCCL} & Kernel fuzzy competitive learning~\cite{mizutani2005kernel}; features \\
\rowcolor{gray!10}
& \texttt{SKFCM} & Spatial kernel FCM; pixels + shape \\
\rowcolor{gray!10}
& \texttt{SoftKSC} & Soft kernel spectral~\cite{langone2013soft}; labelled + unlabelled \\
\rowcolor{gray!10}
& \texttt{SoftDBSCANGM} & Soft DBSCAN-GM~\cite{smiti2016fuzzy}; features \\

Mixture and
& \texttt{MBMM} & Multivariate beta mixture~\cite{hsu2024multivariate}; features in $(0,1)$ \\
evidential
& \texttt{BGMM} & Beta-Gaussian mixture~\cite{dai2008bgmm}; two views \\
& \texttt{ECM} & Evidential c-means~\cite{masson2008ecm}; features \\
& \texttt{EVCLUS} & Evidential clustering of proximities~\cite{denoeux2004evclus}; features or dissimilarities \\

\rowcolor{gray!10}
Graph and
& \texttt{BIGCLAM} & BigCLAM~\cite{yang2013overlapping}; adjacency \\
\rowcolor{gray!10}
community
& \texttt{BayesianNMF} & Bayesian NMF~\cite{psorakis2011overlapping}; adjacency \\
\rowcolor{gray!10}
& \texttt{MMSB} & Mixed-membership blockmodel~\cite{airoldi2008mixed}; adjacency \\
\rowcolor{gray!10}
& \texttt{CDCGS} & Gumbel-softmax community detection~\cite{acharya2020community,jang2017categorical}; adjacency \\
\rowcolor{gray!10}
& \texttt{NOCD} & Overlapping community GNN~\cite{shchur2019overlapping}; sparse graph + features \\
\rowcolor{gray!10}
& \texttt{DMoN} & Deep modularity networks~\cite{tsitsulin2023graph}; features + graph \\

Ensemble
& \texttt{SCSPA} & Soft CSPA consensus~\cite{punera2008consensus}; partitions \\
(consensus)
& \texttt{SHBGF} & Soft HBGF consensus~\cite{punera2008consensus}; partitions \\
& \texttt{SMCLA} & Soft MCLA consensus~\cite{punera2008consensus}; partitions \\

\rowcolor{gray!10}
Advanced
& \texttt{AFCM} & Adaptive FCM with graph embedding~\cite{chen2024adaptive}; features \\
\rowcolor{gray!10}
fuzzy
& \texttt{AFCMSimple} & The same objective without the graph term; features \\
\rowcolor{gray!10}
& \texttt{AFCMAdaptive} & Adaptive FCM with bias field~\cite{pham1999adaptive}; grayscale image \\
\rowcolor{gray!10}
& \texttt{RPFKM} & Robust projected fuzzy $k$-means~\cite{zhao2022improving}; features \\
\rowcolor{gray!10}
& \texttt{RDFKC} & Robust deep fuzzy $k$-means~\cite{wu2024robust}; features + autoencoder \\
\rowcolor{gray!10}
& \texttt{FeMIFuzzy} & Federated fuzzy clustering~\cite{ngo2023federated} with Sammon projection~\cite{sammon1969nonlinear}; client matrices \\
\rowcolor{gray!10}
& \texttt{SFCMEP} & Semi-supervised FCM~\cite{xu2024semi}; features + partial labels \\
\rowcolor{gray!10}
& \texttt{PFCM} & Possibilistic FCM~\cite{pal2005possibilistic}; features \\

Specialized
& \texttt{ENTROPYFCM} & Entropy-regularised c-means~\cite{gupta2019fuzzy}; features \\
& \texttt{CAFHFCM} & Centroid auto-fused hierarchical FCM~\cite{lin2020centroid}; features \\
& \texttt{CAFCM} & Collaborative annealing FCM~\cite{li2023soft}; features \\
& \texttt{FCC} & Fuzzy colour clustering~\cite{kim2024fuzzy}; features \\
& \texttt{WBSC} & Word-based soft clustering~\cite{lin2001word}; documents \\
& \texttt{SISC} & Similarity-based soft clustering~\cite{lin2001similarity}; documents \\
& \texttt{KMART} & Modified fuzzy ART~\cite{kondadadi2002modified}; documents \\

\bottomrule
\end{tabularx}
\caption{The 42 algorithms SCPP exports. \emph{Input} is the first argument
of \texttt{fit}. Every row is fitted by the conformance suite on every commit.}
\label{tab:scpp_algorithm_families}
\end{table*}

\section{Benchmark Data Set Collection}
\label{app:dataset}

SCPP provides a curated suite of 20 benchmark data sets spanning
\emph{real-world}, \emph{synthetic}, and \emph{OpenML} data. All data sets are
exposed through a common API that returns a feature matrix
$X \in \mathbb{R}^{n \times d}$ and ground-truth labels $y$. The labels are
used only for external evaluation (ARI and NMI), not for fitting.

Four of the five real-world data sets ship with scikit-learn and the six
synthetic data sets are generated locally, so ten of the eleven require no
network access; \texttt{olivetti\_faces} is downloaded on first use through
\texttt{fetch\_olivetti\_faces} and cached thereafter. The remaining
nine data sets are retrieved from OpenML on first use through
\texttt{fetch\_openml} at \texttt{version=1}; consequently, their availability
depends on the
network and the corresponding OpenML versions. OpenML may also return string
class labels, which does not affect ARI or NMI because both metrics are
invariant to label encoding (see Tables~\ref{tab:real-datasets}, \ref{tab:synthetic-datasets}, and \ref{tab:openml-datasets}). 

\begin{table}[t]
\centering
\small
\begin{tabular}{lrrrl}
\toprule
\textbf{Data Set} & $n$ & $d$ & $K$ & \textbf{Source} \\
\midrule
\texttt{iris}           & 150     & 4    & 3  & UCI / scikit-learn \\
\texttt{wine}           & 178     & 13   & 3  & UCI / scikit-learn \\
\texttt{digits}         & 1{,}797 & 64   & 10 & NIST / scikit-learn \\
\texttt{breast\_cancer} & 569     & 30   & 2  & UCI / scikit-learn \\
\texttt{olivetti\_faces} & 400    & 4{,}096 & 40 & AT\&T / scikit-learn \\
\bottomrule
\end{tabular}
\caption{Real-world benchmark data sets.}
\label{tab:real-datasets}
\end{table}

\begin{table}[t]
\centering
\small
\begin{tabular}{lrrlp{4.5cm}}
\toprule
\textbf{Data Set} & $n$ & $d$ & $K$ & \textbf{Characteristics} \\
\midrule
\texttt{blobs} &
1{,}000 & 2 & 5 & Isotropic Gaussians \\
\texttt{moons} &
1{,}000 & 2 & 2 & Non-linear, noise $=0.05$ \\
\texttt{circles} &
1{,}000 & 2 & 2 & Concentric, factor $=0.5$ \\
\texttt{anisotropic\_blobs} &
1{,}500 & 2 & 4 & Linearly transformed Gaussians \\
\texttt{varied\_blobs} &
1{,}500 & 2 & 4 & Unequal cluster variances \\
\texttt{high\_dimensional\_blobs} &
3{,}000 & 100 & 10 & High-dimensional stress test \\
\bottomrule
\end{tabular}
\caption{Synthetic benchmark data sets. All are produced by the
\texttt{scikit-learn} generators at \texttt{random\_state=42};
\texttt{circles} also uses \texttt{noise}~$=0.05$.}
\label{tab:synthetic-datasets}
\end{table}

\begin{table}[t]
\centering
\small
\begin{tabular}{lrrr}
\toprule
\textbf{Data Set} & $n$ & $d$ & $K$ \\
\midrule
\texttt{glass}     & 214    & 9  & 6  \\
\texttt{vehicle}   & 846    & 18 & 4  \\
\texttt{ecoli}     & 336    & 7  & 8  \\
\texttt{yeast}     & 1{,}484 & 8  & 10 \\
\texttt{segment}   & 2{,}310 & 19 & 7  \\
\texttt{satimage}  & 6{,}430 & 36 & 6  \\
\texttt{letter}    & 20{,}000 & 16 & 26 \\
\texttt{pendigits} & 10{,}992 & 16 & 10 \\
\texttt{optdigits} & 5{,}620 & 64 & 10 \\
\bottomrule
\end{tabular}
\caption{OpenML benchmark data sets.}
\label{tab:openml-datasets}
\end{table}

All data sets can be loaded individually or by group through the same API:

\begin{lstlisting}[language=Python, caption={Loading benchmark data sets through the unified API.}]
from soft_clustering.benchmarking import (
    get_dataset, benchmark_suite, available_datasets
)

X, y = get_dataset("iris")

real_suite = benchmark_suite("real")
synth_suite = benchmark_suite("synthetic")
openml_suite = benchmark_suite("openml")

print(available_datasets())
\end{lstlisting}

\section{Soft Clustering Evaluation Framework}
\label{app:metrics}

SCPP provides twelve cluster-validity measures: five applied to defuzzified
labels (Silhouette, Calinski--Harabasz, Davies--Bouldin, ARI, and NMI) and
seven that operate directly on the membership matrix $U$. The latter are
Partition Coefficient (PC), Modified Partition Coefficient (MPC), Partition
Entropy (PE), Fuzzy Hypervolume (FHV), Xie--Beni (XB), Fuzzy Compactness (FC),
and mean pairwise prototype separation.

\paragraph{Soft metrics.}
Let $u_{ik}$ denote the membership of sample $k$ in cluster $i$, $v_i$ its
prototype, and $m$ the fuzzifier. The implemented measures are
\begin{align*}
\mathrm{PC} &=
\frac{1}{n}\sum_{i,k} u_{ik}^{2},
&
\mathrm{MPC} &=
\frac{\mathrm{PC}-1/K}{1-1/K},
&
\mathrm{PE} &=
-\frac{1}{n}\sum_{i,k} u_{ik}\log u_{ik},\\
\mathrm{FHV} &=
\sum_i \sqrt{\det(F_i)},
&
\mathrm{XB} &=
\frac{\sum_{i,k}u_{ik}^{m}\lVert x_k-v_i\rVert^2}
{n\min_{i\neq j}\lVert v_i-v_j\rVert^2},
&
\mathrm{FC} &=
\sum_{i,k}u_{ik}^{m}\lVert x_k-v_i\rVert^2,
\end{align*}
where
\[
F_i =
\frac{\sum_k u_{ik}^{m}(x_k-v_i)(x_k-v_i)^\top}
{\sum_k u_{ik}^{m}}
\]
is the fuzzy covariance of cluster $i$. PC and PE follow
\citet{bezdek1981pattern}, MPC \citet{dave1996validating}, XB
\citet{xie1991validity}, and FHV \citet{gath1989fuzzy}. FHV measures the
aggregate volume occupied by the clusters and is therefore meaningful only
for partitions of the same feature space.

\paragraph{Interpretation and applicability.}
PC, MPC, and PE measure partition \emph{crispness}, not clustering quality.
For example, a hard partition achieves $\mathrm{PC}=1$ and
$\mathrm{PE}=0$ regardless of whether the resulting clusters are meaningful.
We therefore report these measures for diagnostic purposes but do not use
them to rank algorithms. They are also defined only for partition-constrained
memberships, i.e., when each row of $U$ sums to one. Seven estimators (PCM, BIGCLAM, BayesianNMF, NOCD, WBSC, SISC and KMART) produce unnormalised
typicalities or affiliations, for which
these measures are not directly comparable. SCPP consequently returns
\texttt{nan} when a metric is not applicable, while retaining the same metric
schema for every estimator. The four prototype-based indices likewise return
\texttt{nan} for estimators that expose no \texttt{centers\_}, so a row of the
released benchmark data may legitimately report fewer than twelve numbers.

The remaining measures capture different aspects of clustering structure:
FHV measures cluster volume, XB balances within-cluster compactness against
prototype separation, FC measures fuzzy within-cluster dispersion, and
prototype separation measures the average distance between cluster
prototypes. These metrics are complementary rather than interchangeable and
are therefore reported individually.

\begin{lstlisting}[language=Python, caption={Computing soft and defuzzified clustering metrics.}]
from soft_clustering.benchmarking import (
    soft_clustering_metrics,
    clustering_metrics,
)

U = model.memberships_
centers = model.centers_

soft = soft_clustering_metrics(
    X, U, centers=centers, m=2.0,
    partition_constrained=model._partition_constrained,
)

hard = clustering_metrics(
    X, np.argmax(U, axis=1), y_true=y
)
\end{lstlisting}

\section{Cross-Family Comparison}
\label{app:performance}

The common protocol enables a comparison across methodological families that
would otherwise require separate implementations and evaluation pipelines.
We therefore evaluate every feature-matrix estimator in SCPP on the eleven
real and synthetic data sets of the shipped suite, using the same fitting
procedure and metrics.

\paragraph{Protocol.}
All 20 feature-matrix estimators are evaluated on all 11 data sets, yielding
220 fits. The number of clusters is set to the number of ground-truth classes,
except for SCM and SoftDBSCANGM, which determine $K$ themselves. Data sets with
more than 2{,}000 samples are subsampled using a fixed seed to prevent a
single data set from dominating runtime. Each fit runs in an isolated
subprocess with a 60\,s timeout.
Metrics follow Appendix~\ref{app:metrics}. The complete benchmark is
reproducible with:

\begin{lstlisting}[style=bashstyle, caption={Reproducing the cross-family benchmark.}]
python benchmarks/run_main_benchmark.py
python benchmarks/run_main_benchmark.py --list-modalities
\end{lstlisting}

\paragraph{Scope.}
The benchmark covers only estimators that consume a feature matrix. The
remaining 22 estimators require other inputs, including graphs, documents,
partitions, images, multiple views, client matrices, partial labels, or
autoencoder representations. Combining these methods with feature-matrix
estimators would require incompatible input representations and would not
constitute a meaningful comparison. The \texttt{--list-modalities} option
enumerates all 42 estimators by input type.

\paragraph{Results.}
Of the 220 fits, 188 completed successfully, 17 produced a single-cluster
partition, 12 timed out, 3 exhausted memory. All fifteen timeouts and memory failures occurred on
\texttt{olivetti\_faces} ($n=400$, $d=4{,}096$, $K=40$). The 60\,s budget
covers the fit and the subsequent metric computation together, and at these
dimensions the fuzzy hypervolume forms a $(K, d, d)$ array of
$40 \times 4096^2$ doubles, so a timeout on this data set does not by itself
locate the cost in the estimator. No run is dropped: every failure is recorded
with its status and reason in the released per-run data.
Table~\ref{tab:main-benchmark} aggregates the per-run records by estimator.

\begin{table}[t]
\centering
\small
\caption{Cross-family comparison: all 20 feature-matrix estimators over the 11
offline datasets of the shipped suite, 220 fits under one protocol.
\emph{Compl.} counts fits returning a non-degenerate partition, \emph{Err.}
timeouts, memory kills and exceptions; no run is dropped. ARI, NMI and
$\bar{t}$ are means over completed runs. The fuzziness indices (PC, MPC, PE)
are released per run but omitted here: they measure crispness, not quality,
so ranking by them would reward a fuzzy method for degenerating to a hard
partition.}
\label{tab:main-benchmark}
\begin{tabular}{lrrrrrr}
\toprule
Algorithm & Datasets & Compl. & Err. & ARI $\uparrow$ & NMI $\uparrow$ & $\bar{t}$ (ms) \\
\midrule
GMM & 11 & 10 & 0 & 0.661 & 0.673 & 57.8 \\
GathGeva & 11 & 10 & 0 & 0.601 & 0.643 & 3839.7 \\
RPFKM & 11 & 8 & 1 & 0.601 & 0.636 & 503.6 \\
CAFCM & 11 & 10 & 0 & 0.589 & 0.614 & 113.5 \\
GK & 11 & 10 & 0 & 0.572 & 0.588 & 227.9 \\
FCM & 11 & 10 & 0 & 0.525 & 0.569 & 11.9 \\
PFCM & 11 & 10 & 0 & 0.497 & 0.539 & 28.6 \\
FCC & 11 & 4 & 6 & 0.486 & 0.577 & 1873.2 \\
ECM & 11 & 10 & 0 & 0.478 & 0.537 & 531.2 \\
ENTROPYFCM & 11 & 10 & 0 & 0.475 & 0.534 & 17.1 \\
AFCMSimple & 11 & 11 & 0 & 0.453 & 0.506 & 185.0 \\
CAFHFCM & 11 & 10 & 0 & 0.389 & 0.430 & 9.9 \\
PCM & 11 & 10 & 0 & 0.385 & 0.476 & 62.4 \\
KFCM & 11 & 10 & 0 & 0.353 & 0.365 & 9.5 \\
SCM & 11 & 10 & 1 & 0.322 & 0.544 & 21.3 \\
EVCLUS & 11 & 11 & 0 & 0.317 & 0.399 & 332.0 \\
SoftDBSCANGM & 11 & 8 & 2 & 0.303 & 0.540 & 2216.8 \\
RoughKMeans & 11 & 11 & 0 & 0.137 & 0.294 & 802.7 \\
AFCM & 11 & 11 & 0 & 0.101 & 0.174 & 9028.0 \\
KFCCL & 11 & 4 & 7 & -0.002 & 0.109 & 341.9 \\
\bottomrule
\end{tabular}
\end{table}

\paragraph{Interpretation.}
The results are intended to characterize cross-family behavior under a common
protocol, not to establish a universal ranking of algorithms. External
metrics such as ARI and NMI measure agreement with the supplied class labels
and therefore favor methods whose inductive biases align with those labels.
Likewise, degenerate results can reflect default hyperparameters rather than
an implementation defect; for example, KFCCL produced a single-cluster
partition on 7 of the 11 data sets under its default learning-rate schedule.
The benchmark should therefore be read as a reproducible characterization of
performance, applicability, and practical failure modes rather than a
leaderboard.

All per-run results, including the twelve metrics, fitted cluster count, and
failure reason, are released in
\texttt{benchmarks/results/main\_benchmark.csv}; the corresponding execution
environment is recorded in \texttt{main\_benchmark\_env.json}.

\section{Agreement with Independent Implementations}
\label{app:external}

Internal tests can verify consistency without detecting a self-consistently
wrong implementation. We therefore validate SCPP against independent
implementations, known limiting cases, and properties implied by the
algorithms themselves.

\paragraph{External agreement.}
Four estimators have an independently maintained implementation of the same
algorithm and are compared against it directly: fuzzy c-means with
\texttt{scikit-fuzzy}, Gaussian mixtures with \texttt{scikit-learn},
evidential c-means with \texttt{evclust}, and LDA with \texttt{scikit-learn}.
Across the comparisons in which the two membership matrices encode the same
quantity, the largest disagreement is $6.2\times10^{-10}$ and all hard
partitions agree. For ECM, the two implementations represent residual
evidential mass differently, so we compare prototypes and partitions rather
than force an invalid element-wise comparison. All comparisons use isotropic
Gaussian clusters at $d = 8$, $K = 3$ with well-separated centres, so they
establish that the two implementations reach the same solution on that
family, not across the input space. The comparison in Table~\ref{tab:external-baselines} and the checks described
above are reproduced with:

\begin{lstlisting}[style=bashstyle, caption={Reproducing the third-party agreement results.}]
python benchmarks/run_external_baselines.py --sizes 200 1000 --repeats 3
pytest tests/test_external_agreement.py -v
\end{lstlisting}

\begin{table}[t]
\centering
\caption{SCPP against independent implementations of the same algorithms.
\emph{Max.\ diff.} is the largest absolute difference between membership
matrices after resolving arbitrary cluster labelling; \emph{Proto.\ gap} is
the largest distance from an SCPP prototype to its nearest reference
prototype; ARI compares the hard partitions. ``n/a'' indicates that the
quantity is not defined for both implementations. For ECM, \texttt{evclust}
places residual evidential mass on the frame $\Omega$, whereas SCPP places it
on a noise cluster; hence their membership matrices are not directly
differenced. LDA has no prototypes. Runtimes are the minimum of three timed
fits after one warm-up, on the machine of Appendix~\ref{app:optimization}.}
\label{tab:external-baselines}

\scriptsize
\setlength{\tabcolsep}{2.5pt}
\renewcommand{\arraystretch}{0.95}

\begin{tabular*}{\columnwidth}{@{\extracolsep{\fill}}llrrrrrr@{}}
\toprule
Algorithm & Reference & $n$ & Max.\ diff. & Proto.\ gap & ARI &
SCPP (ms) & Ref.\ (ms) \\
\midrule
FCM & scikit-fuzzy cmeans & 200
& 6.2e-10 & 3.1e-10 & 1.0000 & 0.5 & 0.6 \\

GMM & scikit-learn GaussianMixture & 200
& 1.0e-12 & 2.3e-13 & 1.0000 & 4.2 & 1.4 \\

ECM & evclust ecm & 200
& n/a & 2.2e-02 & 1.0000 & 5.9 & 22.8 \\

LDA & scikit-learn LDA & 200
& 1.1e-10 & --- & 1.0000 & 937.8 & 346.2 \\

FCM & scikit-fuzzy cmeans & 1,000
& 2.2e-10 & 8.9e-11 & 1.0000 & 1.8 & 1.7 \\

GMM & scikit-learn GaussianMixture & 1,000
& 1.0e-12 & 1.2e-14 & 1.0000 & 7.9 & 2.5 \\

ECM & evclust ecm & 1,000
& n/a & 1.4e-02 & 1.0000 & 21.0 & 108.5 \\
LDA & scikit-learn LDA & 1,000 & 8.1e-13 & --- & 1.0000 & 4683.9 & 1593.2 \\
\bottomrule
\end{tabular*}
\end{table}

\section{Testing}\label{Testing}

SCPP has \textbf{1{,}206 tests across 48 modules} with \textbf{95\% line
coverage}. The tests combine estimator-specific checks with protocol,
edge-case, external-agreement, and optimization-equivalence suites.

\subsection{Organisation}

The 42 estimator modules test output shapes, invariants, parameter validation,
convergence, and planted-structure recovery where applicable. The
cross-cutting suites test the common protocol (\texttt{test\_protocol.py},
522 tests), degenerate inputs (\texttt{test\_edge\_cases.py}, 130 tests),
third-party agreement (\texttt{test\_external\_agreement.py}, 32 tests),
optimization
equivalence (\texttt{test\_optimization\_equivalence.py}, 36 tests), and the
benchmarking and public API infrastructure
(\texttt{test\_benchmarking.py}, 109 tests).

\subsection{Conformance and Edge Cases}

Every exported estimator is constructed and fitted on a valid input of its
modality. The conformance suite verifies that \texttt{fit} returns
\texttt{self}; \texttt{memberships\_} has shape $(n,K)$ and is non-negative;
partition-constrained estimators produce rows summing to one;
$\texttt{labels\_}=\arg\max U$; \texttt{n\_clusters} matches the membership
dimension; \texttt{predict()} fails before fitting; \texttt{predict(X)}
either produces valid predictions or raises; parameters round-trip through
\texttt{sklearn.clone}; and, for the fifteen estimators whose conformance case
supplies one, a fixed \texttt{random\_state} reproduces the
memberships. The suite is tied to the public registry and permits no
exclusions, a dedicated test asserts that the exclusion list is empty, so every exported estimator must satisfy the same contract.

Edge-case tests apply the same rule: an estimator must either return a valid
result or raise explicitly. They cover the seventeen feature-matrix
estimators that the harness can construct from a cluster count and three of
the graph estimators, over duplicate and identical points,
singular covariances, $K>n$, $K=1$, extreme fuzzifiers, \texttt{float32} and
non-contiguous inputs, non-finite values, and empty and complete graphs. These
tests exposed a numerical instability in nine modules: the standard
fuzzy-membership update can overflow when a sample coincides with a prototype
or $m$ approaches one. SCPP now evaluates the equivalent expression in a
numerically stable form, factoring out the minimum distance before
exponentiation and using log-space evaluation where required.

\begin{table}[t]
\centering
\rowcolors{2}{white}{gray!5}
\begin{tabular}{p{5.2cm} p{9.5cm}}
\toprule
\textbf{Task} & \textbf{Command} \\
\midrule
Install development dependencies &
\texttt{pip install -e ".[dev,deep,baselines]"} \\
Run the complete suite & \texttt{pytest} \\
Coverage report &
\texttt{pytest --cov=soft\_clustering --cov-report=term-missing} \\
Conformance only & \texttt{pytest tests/test\_protocol.py -v} \\
Edge cases only & \texttt{pytest tests/test\_edge\_cases.py -v} \\
Third-party agreement &
\texttt{pytest tests/test\_external\_agreement.py -v} \\
Optimization equivalence &
\texttt{pytest tests/test\_optimization\_equivalence.py -v} \\
Regenerate paper statistics &
\texttt{python tools/paper\_stats.py --with-coverage} \\
\bottomrule
\end{tabular}
\caption{Commands for running the SCPP test suite.}
\label{tab:testing_commands}
\end{table}

\section{Case Study: Pluggable Clustering in a Migration Pipeline}
\label{appendix:mo2om}

To test the protocol beyond SCPP itself, we integrated it into Mo2oM, a
monolith-to-microservice migration pipeline
\citep{ziabakhsh2025extracting}. Mo2oM builds structural and semantic
similarity graphs over software classes, clusters them into candidate
services, and combines the resulting memberships. Its original clustering
stage was tied to NOCD, making comparison with alternative methods require
changes to the pipeline.

\begin{lstlisting}[language=Python, caption={Original clustering stage: the method is hard-wired into the pipeline.}]
unixcoderMem = NOCD(adjacencyMatrix, semanticSim, n_clusters)
structuralMem = NOCD(adjacencyMatrix, structuralSim, n_clusters)
combinedMem = alpha * unixcoderMem + (1 - alpha) * structuralMem
\end{lstlisting}

With SCPP, the clustering method becomes a replaceable component while each
estimator retains its native input requirements:

\begin{lstlisting}[language=Python, caption={SCPP integration: the clustering method becomes a parameter.}]
import soft_clustering as sc

MODELS = {
    "NOCD":        lambda k, n: sc.NOCD(random_state=42, max_epochs=10),
    "BayesianNMF": lambda k, n: sc.BayesianNMF(n_clusters=k),
    "BIGCLAM":     lambda k, n: sc.BIGCLAM(n_nodes=n, n_clusters=k),
}

model = MODELS[clustering_method](n_clusters, n_nodes)
model.fit(adjacencyMatrix)
combinedMem = alpha * model.memberships_ + (1 - alpha) * structuralMem
\end{lstlisting}

These three estimators are interchangeable behind a single \texttt{fit} call
because all three consume an adjacency matrix alone. The case study therefore demonstrates the practical purpose of the protocol
within one input modality: graph clustering algorithms can be substituted
without rewriting the surrounding evaluation pipeline. The Mo2oM pipeline
itself is external to this repository, so the migration-quality measurements
it produces are outside the reproducibility guarantees of
Section~\ref{sec:quality}.

\section{Implementation Optimization Study}\label{app:optimization}

Rather than assume the implementations are computationally efficient, we
measure them under a common protocol, and rewrite the ones measurement says
are worth rewriting. This appendix gives the procedure, its coverage, the
five cases, and the complete results. Every artifact, raw records,
consolidated results, tables, figures and the commands that produce them,
is in the repository under \texttt{optimization/}.

\paragraph{What the baseline is, and what it is not.}
The comparison here is between SCPP's current implementations and SCPP's own
pre-optimization implementations, preserved as an importable reference build.
It answers ``did the rewrite help, and did it change the answer'', which is
the question a maintainer must answer before shipping a rewrite. It does
\emph{not} answer ``is SCPP faster than the alternatives''; the speedups
below are relative to code we wrote, and a large speedup indicates a
correspondingly slow starting point. Appendix~\ref{app:external} is the
comparison against third-party implementations, and is where a reader should
look for that question.

\subsection{Protocol}

All 42 exported algorithms were audited statically for loop structure and
numerical call sites. Of these, 31 can be constructed by a shared input
harness and were baseline-timed at $n = 200$ and 30 completed.
The 10 slowest were profiled at the function level. Where profiling
identified a specific bottleneck, the implementation was reworked and the
original preserved as a separately importable reference, and the two were
compared for numerical equivalence before any speedup was reported.

An optimization counted as equivalent when the maximum absolute difference
between membership matrices was below $10^{-9}$, hard-label agreement was
exactly $1.0$, the discovered cluster counts matched, and the output
invariants held: memberships finite and non-negative, and
$\text{labels} = \arg\max U$.

\begin{figure}[t]
\centering
\includegraphics[width=0.60\textwidth]{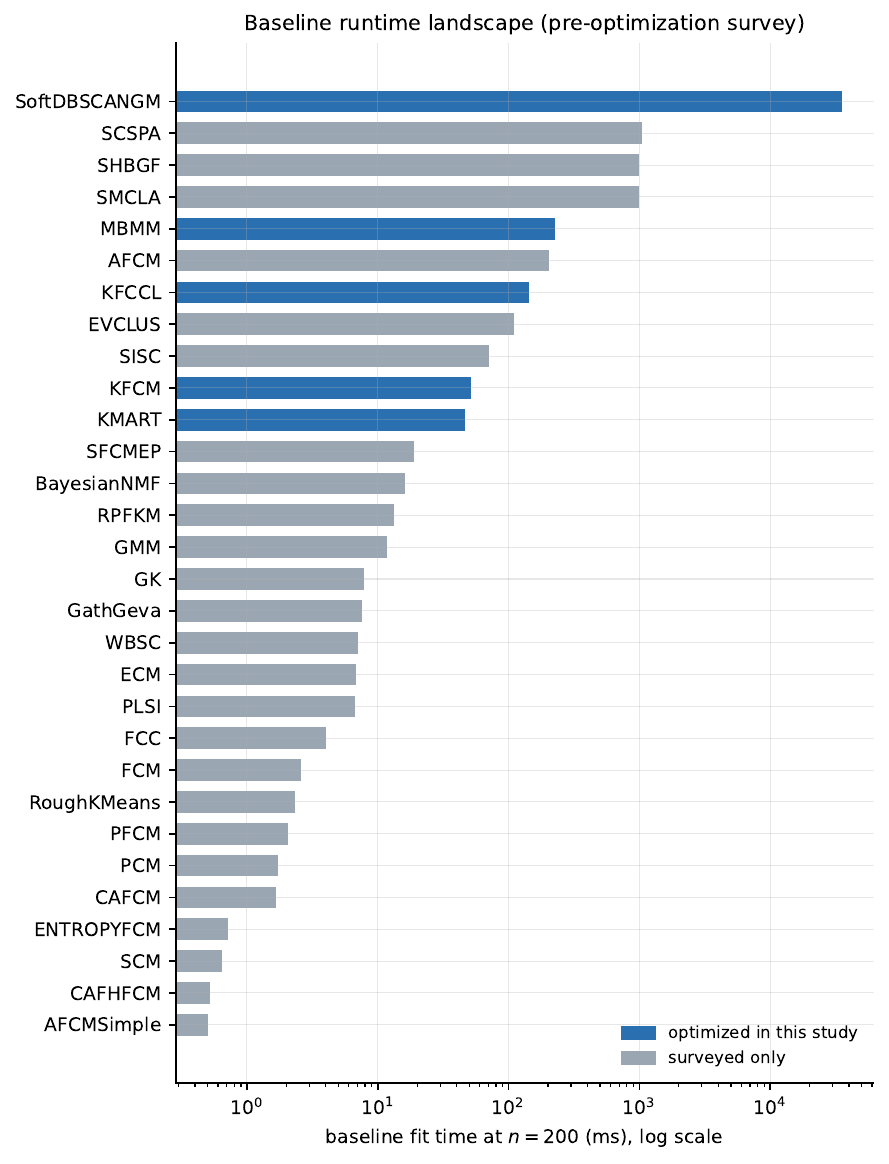}
\caption{Baseline fit time of every algorithm the shared harness can
construct, at $n = 200$ (log scale). Highlighted bars are the algorithms
subsequently optimized. Fifteen fall below $10$\,ms and were left unchanged.
Each algorithm is timed on $n = 200$ inputs of its own modality, feature
matrices, documents, ensembles or a graph, so this is a triage instrument
for locating implementations worth profiling, not a like-for-like ranking.}
\label{fig:opt-landscape}
\end{figure}

\paragraph{The harness is common in protocol, not in workload.}
The 30 baseline-timed algorithms do not all see the same array. They see
$n = 200$ in \emph{their own input modality}: 22 receive a $200 \times 8$
feature matrix, four receive 200 short documents, three receive an ensemble
of three $200 \times 3$ partitions, and one receives a 200-node graph. This
is unavoidable, the algorithms do not accept a common input, but it
means Figure~\ref{fig:opt-landscape} compares work done on different objects
and should be read as a triage instrument for finding implementations worth
profiling, not as a ranking of algorithmic cost.

All measurements are on an Apple M2 (8 cores, macOS 26.0, arm64) with Python
3.13.5, NumPy 2.5.2, SciPy 1.18.0 and scikit-learn 1.9.0.
Each fit runs in a separate subprocess to isolate allocator and import state.
Paired benchmarks use one warm-up fit and three timed repetitions, reporting
the minimum; all repetitions are retained in the released data. Timeouts are
recorded as results, so an algorithm that does not finish is reported as
such.

\begin{table}[t]
\centering
\setlength{\tabcolsep}{8pt}
\renewcommand{\arraystretch}{1.15}
\small
\definecolor{scppbg}{RGB}{220,252,231}
\begin{tabular}{lr}
\toprule
\textbf{Stage} & \textbf{Algorithms} \\
\midrule
\rowcolor{gray!10}
Exported by SCPP & 42 \\
Statically audited & 42 \\
\rowcolor{gray!10}
Constructible by the shared harness & 31 \\
\quad of which baseline-timed successfully & 30 \\
\rowcolor{gray!10}
Profiled at function level & 10 \\
\rowcolor{scppbg}
\textbf{Optimized, verified and benchmarked end to end} & \textbf{5} \\
\midrule
Profiled, bottleneck identified, not yet optimized & 5 \\
Timed and left unchanged & 20 \\
\quad of which complete a 200-sample fit in $<10$\,ms & 15 \\
\quad of which take $10$--$110$\,ms and were not profiled & 5 \\
Input shape outside the shared harness; audited only & 11 \\
\bottomrule
\end{tabular}
\caption{Coverage of the optimization study. The gap between \emph{profiled}
and \emph{optimized} is deliberate: measurement said the rest were not worth
changing.}
\label{tab:opt-coverage}
\end{table}

\subsection{The Baseline Landscape}

Figure~\ref{fig:opt-landscape} plots the baseline fit time of every algorithm
the harness can construct. The distribution is strongly skewed: fifteen
complete a 200-sample fit in under $10$\,ms, where any Python-level
improvement would be irrelevant to total runtime, while SoftDBSCANGM takes
$35.3$\,s on the same sample count, more than three orders of magnitude
above the median of $7.7$\,ms. That gap, not inspection of the source, is what selected the
implementations for profiling.

\subsection{A Recurring Defect Class}\label{app:opt:exact}

Four of the five optimizations addressed the same underlying implementation
pattern: a scalar numerical operation was repeatedly invoked inside a Python
loop even though an equivalent vectorized or batched computation was available.
SoftDBSCANGM called \texttt{scipy.spatial.distance.mahalanobis} for each
$(i,j,t)$ combination; KFCM evaluated a scalar Gaussian kernel $NK$ times per
iteration; KFCCL recomputed a normalized kernel column for each
(iteration, cluster, sample) combination; and KMART called \texttt{np.minimum}
once for each (document, prototype) pair. In each case, the optimization
removed repeated Python-level work without changing the underlying computation:
the optimized expression is algebraically equivalent to the original or differs
only in the order in which the same operations are evaluated.

\paragraph{Why the rewrites are algebraically equivalent. }\textbf{SoftDBSCANGM.} The membership update evaluated
$U_{ij} = 1 / \sum_t (d_{ij}/d_{it})^{p}$ with a triple Python loop, computing
$NK^2$ distances although only $NK$ distinct values exist. Since
\[
\sum_t \left(\frac{d_j}{d_t}\right)^{p} = d_j^{\,p} \sum_t d_t^{-p},
\qquad\text{so}\qquad
\frac{1}{\sum_t (d_j/d_t)^{p}} = \frac{d_j^{-p}}{\sum_t d_t^{-p}},
\]
the update follows from one $N \times K$ distance matrix without ever forming
an $(N,K,K)$ tensor.

Evaluated as written, $d_j^{-p}$ overflows when a sample lies on a prototype,
and $p = 2/(m-1)$ grows without bound as the fuzzifier approaches one, so the
overflow is reached for ordinary data at small $m$. Rescaling the distances
by their per-sample minimum before exponentiating removes it: equivalently,
multiplying numerator and denominator by $\min_t (d_t)^{p}$,
\[
\frac{d_j^{-p}}{\sum_t d_t^{-p}}
= \frac{\big(\min_t d_t / d_j\big)^{p}}
       {\sum_t \big(\min_t d_t / d_t\big)^{p}},
\]
which is the same value with every base in $(0, 1]$ and therefore every power
in $(0, 1]$. This identity is now the library-wide helper described in
Appendix~\ref{Testing}; nine modules previously carried their own unstable
copy of the rule.

\medskip
\textbf{MBMM.} \texttt{scipy.stats.beta.logpdf} was called once per
(component, feature) in both the E-step and the log-likelihood. Substituting
the identity SciPy evaluates internally,
\[
\log \mathrm{Beta}(x; a, b) = \operatorname{xlogy}(a-1, x)
  + \operatorname{xlog1py}(b-1, -x) - \operatorname{betaln}(a, b),
\]
evaluates all features at once through the same \texttt{scipy.special}
primitives, bypassing \texttt{rv\_continuous} dispatch, argument checking and
support masks. The identity holds on the support of the distribution, which
is where the model is defined; outside $(0,1)$ the two expressions differ,
because SciPy's support mask returns $-\infty$ where the direct evaluation
returns \texttt{nan}.

\subsection{Case Studies}

Table~\ref{tab:opt-cases} summarizes the five implementations that were
optimized after profiling identified specific computational bottlenecks. The
following case studies describe the main source of the inefficiency and the
corresponding implementation change.

\begin{table}[t]
\centering
\footnotesize
\setlength{\tabcolsep}{4pt}
\renewcommand{\arraystretch}{1.2}

\begin{tabular}{llp{5.1cm}rr}
\toprule
\textbf{Algorithm} &
\textbf{Family} &
\textbf{Bottleneck} &
\textbf{Speedup} &
\textbf{Max. deviation} \\
\midrule

\rowcolor{gray!10}
SoftDBSCANGM &
Density &
$NK^2$ scalar Mahalanobis evaluations per iteration; $K$ grows with $N$ &
$889$--$4{,}507$ &
$2.2\mathrm{e}{-16}$ \\

KFCM &
Kernel &
$NK$ scalar Gaussian-kernel evaluations per iteration and quadratic
Python work during K-Means++ initialization &
$82$--$546\times$ &
$4.8\mathrm{e}{-15}$ \\

\rowcolor{gray!10}
KFCCL &
Kernel &
Normalized kernel columns recomputed inside the innermost loop &
$27$--$152\times$ &
$1.1\mathrm{e}{-16}$ \\

KMART &
Document &
Scalar reductions for each document--prototype pair and element-wise
\texttt{lil\_matrix} construction &
$10$--$71\times$ &
$0.0$ \\

\rowcolor{gray!10}
MBMM &
Mixture &
Repeated \texttt{rv\_continuous} dispatch and $KD$ scalar calls in the
E-step and log-likelihood computation &
$6$--$17\times$ &
$2.2\mathrm{e}{-14}$ \\

\bottomrule
\end{tabular}
\caption{Optimized SCPP algorithms and identified computational bottlenecks.
Speedup ranges are measured across the sample sizes used in the paired
benchmarks. The original SoftDBSCANGM implementation did not
finish within $600$\,s at $n=240$; the reported upper bound is therefore a
lower bound.}
\label{tab:opt-cases}
\end{table}

\paragraph{SoftDBSCANGM.}
SoftDBSCANGM exhibited the most significant scalability problem in the study.
Because each DBSCAN noise point can become its own component, the number of
components $K$ increases with the number of samples. The original membership
update repeatedly evaluated distances inside a triple loop, giving cubic
growth in the sample count. The measured runtime increased from
$3{,}410$,ms to $26{,}695$,ms when $n$ was doubled, corresponding to a
factor of $7.83 \approx 2^{2.97}$. The optimized implementation computes the
distance matrix once per iteration using a batched \texttt{einsum}, applies
the equivalent closed-form membership update, uses \texttt{searchsorted} for
the initialization mapping, and performs the matrix inversion in a batched
form. The largest temporary is processed in chunks to limit memory use. This
change reduced the measured runtime by up to $4{,}507\times$ and allowed
problem sizes to be processed that the original implementation could not
complete within the measurement timeout.

\paragraph{KFCM.}
The KFCM profile initially showed substantial time spent in
\texttt{typeguard}, which performs runtime type checking for the estimator.
Closer inspection showed that the overhead resulted primarily from the large
number of calls to a scalar Gaussian-kernel helper inside Python loops. The
optimized implementation evaluates the kernel values in a matrix form,
eliminating the repeated scalar calls and their associated dispatch and type
checking overhead. This change produced speedups of $82$--$546\times$ while
leaving runtime type checking enabled. Numerical equivalence also required
preserving the order and number of random draws used during K-Means++
initialization; a regression test verifies that the optimized implementation
leaves the NumPy random state unchanged relative to the reference
implementation.

\paragraph{KFCCL.}
KFCCL repeatedly reconstructed a normalized kernel column inside the innermost
loop, even though the kernel matrix and its normalization terms remain
constant throughout a fit. The optimized implementation computes these
quantities once for the full matrix and replaces the repeated element-wise
operations with matrix operations. This removes loop-invariant computation
from the inner loop and results in speedups between $27$ and $152\times$.

\paragraph{KMART.}
KMART performs a vigilance calculation based on element-wise minima between
input representations and prototypes. The original implementation evaluated
these operations through Python-level loops and incrementally constructed a
\texttt{lil\_matrix}. The optimized version stores the prototypes in a
contiguous array and performs the same operations as a vectorized reduction.
The resulting implementation preserves the operation order and produced a
maximum numerical deviation of $0.0$ in the paired comparisons, with
prototypes agreeing bitwise.

\paragraph{MBMM.}
MBMM repeatedly called \texttt{scipy.stats.beta.logpdf} for individual
component--feature pairs during both the E-step and log-likelihood
calculation. The optimized implementation evaluates the equivalent
log-density expression directly using vectorized \texttt{scipy.sp ecial}
operations, avoiding repeated distribution-level dispatch and validation.
The resulting speedup ranges from $6$ to $17\times$. The speedup decreases
with increasing $n$ because the removed $O(KD)$ function-call overhead becomes
a smaller fraction of the total computation as the sample-dependent work
grows.

\subsection{Runtime Results}

Table~\ref{tab:opt-runtime} reports every paired measurement, while
Table~\ref{tab:opt-summary} summarizes the overall results. Across the 20
paired measurements in which both implementations completed, the median
speedup is $70.4\times$ and the geometric mean is $72.6\times$. The arithmetic
mean is substantially larger ($365.6\times$) because the measurements combine
ordinary constant-factor improvements with two complexity reductions. We
report all three statistics to give a more complete picture of the observed
performance gains.

\begin{table}[t]
\centering
\setlength{\tabcolsep}{10pt}
\renewcommand{\arraystretch}{1.15}
\small
\begin{tabular}{lr}
\toprule
\textbf{Statistic} & \textbf{Runtime speedup} \\
\midrule
Minimum & $6.3\times$ \\
Median & $\mathbf{70.4\times}$ \\
Geometric mean & $\mathbf{72.6\times}$ \\
Arithmetic mean & $365.6\times$ \\
Maximum (both builds complete) & $4{,}506.8\times$ \\
Maximum (reference times out) & $>47{,}947\times$ \\
\bottomrule
\end{tabular}
\caption{Summary of the 20 paired measurements for which both implementations
completed.}
\label{tab:opt-summary}
\end{table}

\begin{table}[t]
\centering
\small
\setlength{\tabcolsep}{5pt}
\renewcommand{\arraystretch}{1.15}

\begin{tabular}{@{}
    l
    r
    r
    r
    r
    r
@{}}
\toprule
\textbf{Algorithm} &
\textbf{$n$} &
\textbf{Original (ms)} &
\textbf{Optimized (ms)} &
\textbf{Speedup} &
\textbf{Reduction} \\
\midrule

\rowcolor{gray!10}
\textbf{KFCCL}
& 100
& 433.8
& 11.9
& $36.5\times$
& 97.3\% \\

\rowcolor{gray!10}
\phantom{\textbf{KFCCL}}
& 200
& 888.1
& 12.8
& $69.6\times$
& 98.6\% \\

\rowcolor{gray!10}
\phantom{\textbf{KFCCL}}
& 400
& 1,873.3
& 13.9
& $134.8\times$
& 99.3\% \\

\rowcolor{gray!10}
\phantom{\textbf{KFCCL}}
& 800
& 3,795.1
& 25.0
& $152.0\times$
& 99.3\% \\

\rowcolor{gray!10}
\phantom{\textbf{KFCCL}}
& 1,600
& 8,406.0
& 317.3
& $26.5\times$
& 96.2\% \\

\addlinespace[2pt]
\midrule
\addlinespace[2pt]

\textbf{KFCM}
& 100
& 102.9
& 1.3
& $81.9\times$
& 98.8\% \\

& 200
& 183.6
& 1.3
& $146.5\times$
& 99.3\% \\

& 400
& 306.5
& 1.5
& $202.7\times$
& 99.5\% \\

& 800
& 495.7
& 1.5
& $330.5\times$
& 99.7\% \\

& 1,600
& 1,239.5
& 2.3
& $545.5\times$
& 99.8\% \\

\addlinespace[2pt]
\midrule
\addlinespace[2pt]

\rowcolor{gray!10}
\textbf{KMART}
& 200
& 120.8
& 12.3
& $9.8\times$
& 89.8\% \\

\rowcolor{gray!10}
\phantom{\textbf{KMART}}
& 600
& 937.5
& 39.6
& $23.7\times$
& 95.8\% \\

\rowcolor{gray!10}
\phantom{\textbf{KMART}}
& 1,200
& 3,577.8
& 86.3
& $41.5\times$
& 97.6\% \\

\rowcolor{gray!10}
\phantom{\textbf{KMART}}
& 2,400
& 13,763.7
& 193.3
& $71.2\times$
& 98.6\% \\

\addlinespace[2pt]
\midrule
\addlinespace[2pt]

\textbf{MBMM}
& 200
& 841.4
& 49.9
& $16.9\times$
& 94.1\% \\

& 600
& 973.8
& 88.0
& $11.1\times$
& 91.0\% \\

& 1,200
& 1,164.5
& 137.9
& $8.4\times$
& 88.2\% \\

& 2,400
& 1,532.1
& 244.2
& $6.3\times$
& 84.1\% \\

\addlinespace[2pt]
\midrule
\addlinespace[2pt]

\rowcolor{gray!10}
\textbf{SoftDBSCANGM}
& 60
& 3,410.1
& 3.8
& $889.4\times$
& 99.9\% \\

\rowcolor{gray!10}
\phantom{\textbf{SoftDBSCANGM}}
& 120
& 26,694.7
& 5.9
& $4{,}506.8\times$
& 100.0\% \\

\rowcolor{gray!10}
\phantom{\textbf{SoftDBSCANGM}}
& 240
& \textit{timeout} ($>600{,}000$)
& 12.5
& $>47{,}947\times$
& $>99.9\%$ \\

\bottomrule
\end{tabular}
\caption{Runtime comparison between the original and optimized
implementations. Times are the minimum of three timed repetitions after one
warm-up fit. Speedup is defined as
$t_{\mathrm{orig}}/t_{\mathrm{opt}}$, and reduction is the corresponding
percentage decrease in runtime.}
\label{tab:opt-runtime}
\end{table}

\begin{figure}[t]
\centering
\includegraphics[width=\textwidth]{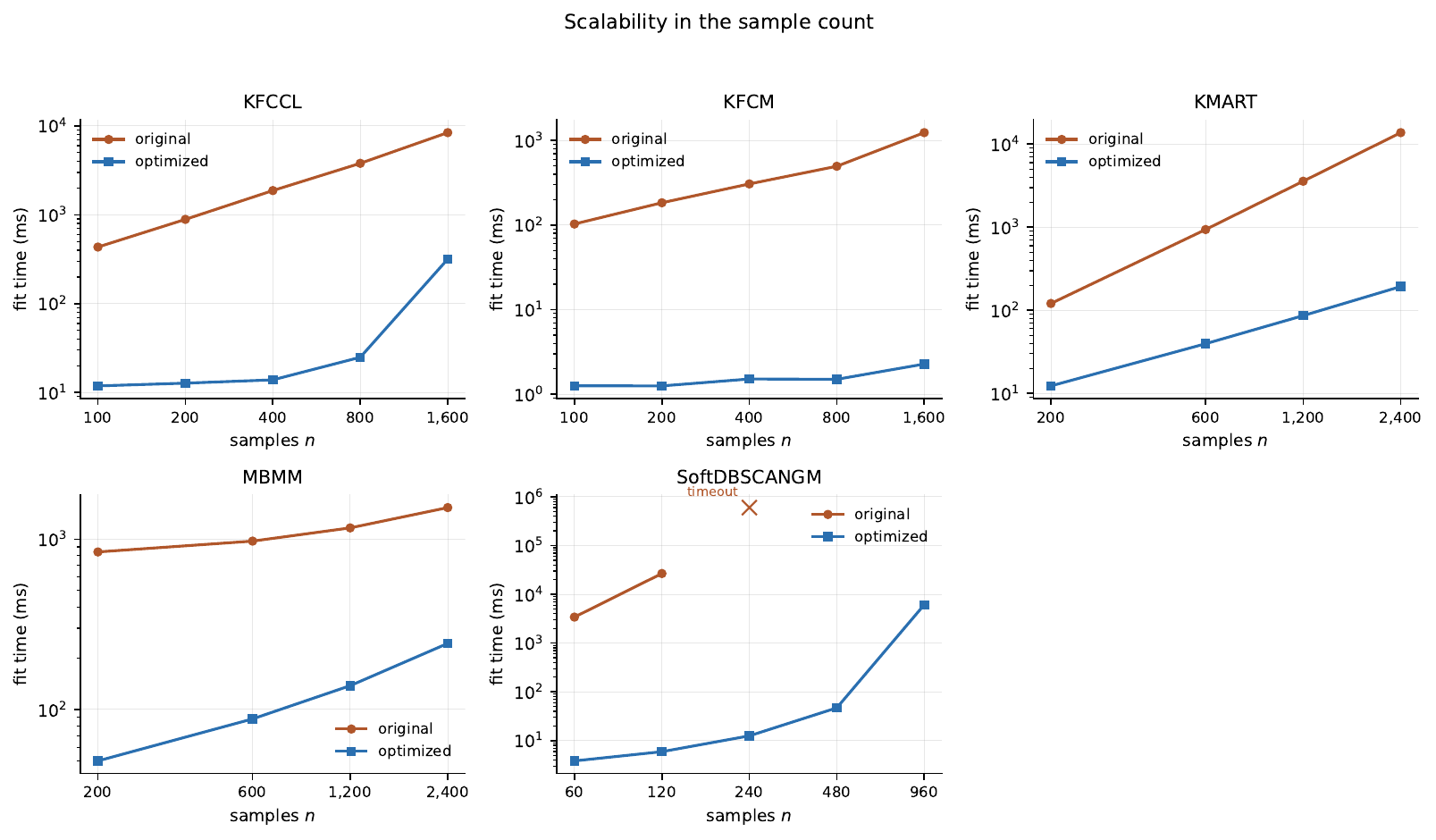}
\caption{Fit time as a function of sample count for the original and optimized
implementations, shown on a log--log scale. The cross denotes a timeout: the
original SoftDBSCANGM implementation does not complete at $n=240$ within
$600$\,s, whereas the optimized implementation completes in $12.5$\,ms.}
\label{fig:opt-scalability}
\end{figure}

Figure~\ref{fig:opt-scalability} shows why a single headline speedup would
not fully describe the results. KFCM and KMART show increasing gains as the
sample size grows, MBMM shows decreasing gains, and SoftDBSCANGM benefits
from a change in computational complexity rather than only a constant-factor
improvement.

\paragraph{A limit reached, not a regression.}
KFCCL is the only non-monotone case: its speedup increases to $152.0\times$ at
$n=800$ and then decreases to $26.5\times$ at $n=1{,}600$. This is not caused
by a change in the number of iterations; both implementations converge in
iteration 65 at that size. Instead, the optimized implementation becomes
memory-bandwidth bound. At $n=1{,}600$, the two $N\times N$ matrices occupy
approximately $41$\,MB and exceed the cache hierarchy. The same effect appears
when timing the underlying BLAS operations in isolation: 195 cluster
iterations of the two matrix--vector products take $6.8$\,ms at $n=800$
($292$\,GB/s effective bandwidth) but $296.9$\,ms at $n=1{,}600$
($26.9$\,GB/s). The latter accounts for essentially all of the measured
$317.3$\,ms runtime. Thus, the optimized implementation has reached the
hardware limit of an algorithm that must repeatedly stream an
$N\times N$ kernel matrix, whereas the original implementation remains
dominated by Python-level call overhead.

\subsection{Correctness}

Performance improvements are only useful if they preserve the original
computation. Each optimized implementation was therefore compared with its
preserved reference on identical inputs across multiple data set sizes and
random seeds. The comparison included membership matrices, hard labels,
cluster prototypes, and the number of discovered clusters. For estimators
using NumPy's global random generator, the generator was seeded immediately
before each fit so that both implementations started from the same state.

Across the 42 comparisons, the maximum absolute difference between membership
matrices was $2.2\times10^{-14}$. Hard labels agreed exactly in every
comparison, and the discovered number of clusters was identical throughout
(Table~\ref{tab:opt-correctness}).

\begin{table}[t]
\centering
\small
\setlength{\tabcolsep}{5pt}
\renewcommand{\arraystretch}{1.15}
\begin{tabular}{lrrrrl}
\toprule
\textbf{Algorithm} &
\textbf{Comparisons} &
\textbf{Max.\ error} &
\textbf{Label agreement} &
\textbf{$K$ match} &
\textbf{Result} \\
\midrule

\rowcolor{gray!10}
KFCCL
& 9
& $1.11\times10^{-16}$
& 1.0000
& Yes
& Equivalent \\

KFCM
& 9
& $4.77\times10^{-15}$
& 1.0000
& Yes
& Equivalent \\

\rowcolor{gray!10}
KMART
& 9
& $\mathbf{0.00}$
& 1.0000
& Yes
& \textbf{Exact} \\

MBMM
& 6
& $2.16\times10^{-14}$
& 1.0000
& Yes
& Equivalent \\

\rowcolor{gray!10}
SoftDBSCANGM
& 9
& $2.22\times10^{-16}$
& 1.0000
& Yes
& Equivalent \\

\midrule

\textbf{Total}
& \textbf{42}
& $\mathbf{2.2\times10^{-14}}$
& \textbf{1.0000}
& \textbf{Yes}
& \textbf{Equivalent} \\

\bottomrule
\end{tabular}
\caption{Correctness comparison between the optimized implementations and
their preserved references. Maximum absolute error is measured over the
membership matrix, and label agreement is the fraction of samples assigned
to the same hard cluster.}
\label{tab:opt-correctness}
\end{table}


These checks are also part of the software test suite rather than being
performed only for the reported experiments. Thirty-six equivalence tests
fit each optimized estimator alongside its preserved reference and run in
continuous integration. A future change that alters the computed result
therefore causes the corresponding CI checks to fail. The tests are skipped
when the reference implementation is unavailable, such as when testing an
installed wheel.

\subsection{Memory}

Table~\ref{tab:opt-memory} gives \emph{every} paired memory measurement, all
twenty of them. Nine are memory-neutral: KFCCL and KMART already materialise
the dominant structure in both builds, an $N \times N$ kernel matrix and a
document-term matrix respectively, so vectorising the work performed on
that array adds nothing ($+3.7\%$ to $-0.1\%$).

The remaining eleven, KFCM, MBMM and SoftDBSCANGM, \textbf{increase}
peak traced allocation, because replacing scalar loops with array operations
materialises intermediates the loops never formed. The relative figures are
large, up to $-896\%$, and they are largest exactly where the absolute
footprint is smallest: KFCM rises from $0.02$ to $0.07$\,MB at $n = 100$, and
SoftDBSCANGM from $0.45$ to $4.45$\,MB at $n = 120$. Chunking bounds the
growth in SoftDBSCANGM. To state the outcome as a fact rather than as a
non-claim: \textbf{three of the five optimizations cost memory and two are
neutral to within $4\%$; none produces a saving of any consequence}. The
speedups are bought with a modest memory increase.

Peak traced allocation is measured with \texttt{tracemalloc} around
\texttt{fit} only. It counts Python-level allocation and therefore understates
buffers held inside NumPy or BLAS; it is consistent between the two builds,
which is what a paired comparison needs, but the absolute figures are not
process resident set size. We did not use resident set size because it is too
noisy for sub-second fits.

\begin{table}[t]
\centering
\small
\setlength{\tabcolsep}{8pt}
\renewcommand{\arraystretch}{1.1}
\begin{tabular}{lrrrr}
\toprule
Algorithm & $n$ & Original (MB) & Optimized (MB) & Change \\
\midrule
KFCCL & 100 & 0.29 & 0.28 & +3.7\% \\
KFCCL & 200 & 0.92 & 0.92 & -0.1\% \\
KFCCL & 400 & 3.66 & 3.67 & -0.1\% \\
KFCCL & 800 & 14.65 & 14.66 & -0.0\% \\
KFCCL & 1,600 & 58.59 & 58.61 & -0.0\% \\
KFCM & 100 & 0.02 & 0.07 & -253.1\% \\
KFCM & 200 & 0.04 & 0.14 & -245.4\% \\
KFCM & 400 & 0.08 & 0.26 & -233.6\% \\
KFCM & 800 & 0.15 & 0.43 & -185.4\% \\
KFCM & 1,600 & 0.26 & 0.86 & -227.4\% \\
KMART & 200 & 0.48 & 0.47 & +0.5\% \\
KMART & 600 & 2.53 & 2.53 & +0.1\% \\
KMART & 1,200 & 8.41 & 8.41 & +0.0\% \\
KMART & 2,400 & 29.82 & 29.82 & +0.0\% \\
MBMM & 200 & 0.03 & 0.07 & -127.3\% \\
MBMM & 600 & 0.08 & 0.21 & -170.7\% \\
MBMM & 1,200 & 0.15 & 0.41 & -168.1\% \\
MBMM & 2,400 & 0.30 & 0.76 & -152.6\% \\
SoftDBSCANGM & 60 & 0.15 & 1.16 & -689.8\% \\
SoftDBSCANGM & 120 & 0.45 & 4.45 & -895.7\% \\
\bottomrule
\end{tabular}
\caption{Peak traced Python allocation, all twenty paired measurements. A
negative change denotes higher usage by the optimized implementation.}
\label{tab:opt-memory}
\end{table}

\subsection{Trade-offs}

\paragraph{Dependencies.}
The optimizations introduce no new dependencies: we use only NumPy and SciPy,
which are already required by SCPP. We considered compiled approaches such as
Cython, Numba, and C++ extensions, but profiling showed that the main
bottlenecks were Python-level call overhead and redundant computation, both of
which could be removed using existing vectorized primitives. We therefore
avoided additional build and packaging complexity without sacrificing the
measured performance gains.

\paragraph{Code complexity.}
The main increase in code size occurs in SoftDBSCANGM, which grows from 115 to
approximately 190 lines because the optimized implementation adds helper
methods and memory-aware chunking. This additional complexity is justified by
the resulting scalability improvement, which makes an otherwise impractical
implementation usable. MBMM remains approximately the same size, while KFCM,
KFCCL, and KMART become more compact by replacing nested Python loops with
vectorized array operations.

\paragraph{Portability and numerical behavior.}
The optimized implementations remain pure NumPy/SciPy code and introduce no
platform-specific or compiled components. BLAS-backed operations can use
different summation orders across platforms, so the equivalence tests use a
conservative tolerance of $10^{-9}$. This is substantially above the largest
observed deviation of $2.2\times10^{-14}$, while still being sufficiently
strict to detect meaningful numerical changes.

\subsection{Reproducibility}

Raw per-measurement records are stored as JSON Lines under
\texttt{optimization/benchmarks/raw/}. The consolidated
\texttt{results.csv} is derived from these records, and every runtime,
memory, correctness and scalability table in this appendix, together with
every figure, is generated from \texttt{results.csv} by a single command.
This ensures that no reported measurement can drift from the underlying
records. The coverage summary of Table~\ref{tab:opt-coverage} and the
bottleneck descriptions of Table~\ref{tab:opt-cases} are written by hand
rather than generated.

\begin{lstlisting}[
    style=bashstyle,
    caption={Regenerating the optimization results and verifying equivalence.},
    label={lst:optimization-reproducibility}
]
# Consolidate raw measurements -> results.csv -> tables + figures
python optimization/analyze.py
python optimization/make_inventory.py

# Verify each optimized estimator against its preserved reference
pytest tests/test_optimization_equivalence.py -v
\end{lstlisting}

\noindent
The complete workflow---static audit, baseline survey, profiling, reference
snapshot, correctness comparison, paired benchmarking, and result
consolidation---is documented in \texttt{optimization/README.md}. The
preserved pre-optimization implementations are stored under
\texttt{optimization/original/scpp\_original/} and remain importable, allowing
the claims reported in this appendix to be independently re-derived from a
fresh checkout.

\bibliography{sample}

\end{document}